# Time Series Forecasting of New Cases and New Deaths Rate for COVID-19 using Deep Learning Methods

Nooshin Ayoobi[1], Danial Sharifrazi[2], Roohallah Alizadehsani[3], Afshin Shoeibi[4,5], Juan M. Gorriz[6], Hossein Moosaei[7], Abbas Khosravi[3], Saeid Nahavandi[3], Abdoulmohammad Gholamzadeh Chofreh[8,*], Feybi Ariani Goni[9], Jiří Jaromír Klemeš[8] and Amir Mosavi[10,11,12,*]

[1]Department of Mathematics, Savitribai Phule Pune University, Pune 411007, India; nooshin.ayoobi@yahoo.com
[2]Department of Computer Engineering, School of Technical and Engineering, Shiraz Branch, Islamic Azad University, Shiraz, Iran; danial.sharifrazi@gmail.com
[3]Institute for Intelligent Systems Research and Innovation, Deakin University, Geelong, VIC 3216, Australia. r.alizadehsani@deakin.edu.au; abbas.khosravi@deakin.edu.au; saeid.nahavandi@deakin.edu.au
[4]Computer Engineering Department, Ferdowsi University of Mashhad, Mashhad, Iran; afshin.shoeibi@gmail.com
[5]Faculty of Electrical and Computer Engineering, Biomedical Data Acquisition Lab, K. N. Toosi University of Technology, Tehran, Iran;
[6]Department of Signal Theory, Networking and Communications, Universidad de Granada, Spain; gorriz@ugr.es
[7]Department of Mathematics, Faculty of Science, University of Bojnord, Iran; moosaei@ub.ac.ir
[8]Sustainable Process Integration Laboratory - SPIL, NETME Centre, Faculty of Mechanical Engineering, Brno University of Technology - VUT Brno, Technická 2896/2, 616 69 Brno, Czech Republic; jiri.klemes@vutbr.cz (J.J.K.)
[9]Department of Management, Faculty of Business and Management, Brno University of Technology - VUT Brno, Kolejní 2906/4, 612 00 Brno, Czech Republic; goni@vutbr.cz
[10]Faculty of Civil Engineering, Technische Universität Dresden, 01069 Dresden, Germany
[11]John von Neumann Faculty of Informatics, Obuda University, 1034 Budapest, Hungary
[12]School of Economics and Business, Norwegian University of Life Sciences, 1430 Ås, Norway
*Corresponding authors:
Abdoulmohammad Gholamzadeh Chofreh, Email: chofreh@fme.vutbr.cz
Amir Mosavi, Email: a.mosavi@ieee.org

**Abstract:** The first known case of Coronavirus disease 2019 (COVID-19) was identified in December 2019. It has spread worldwide, leading to an ongoing pandemic, imposed restrictions and costs to many countries. Predicting the number of new cases and deaths during this period can be a useful step in predicting the costs and facilities required in the future. The purpose of this study is to predict new cases and deaths rate one, three and seven-day ahead during the next 100 days. The motivation for predicting every n days (instead of just every day) is the investigation of the possibility of computational cost reduction and still achieving reasonable performance. Such a scenario may be encountered real-time forecasting of time series. Six different deep learning methods are examined on the data adopted from the WHO website. Three methods are LSTM, Convolutional LSTM, and GRU. The bidirectional extension is then considered for each method to forecast the rate of new cases and new deaths in Australia and Iran countries.

This study is novel as it carries out a comprehensive evaluation of the aforementioned three deep learning methods and their bidirectional extensions to perform prediction on COVID-19 new cases and new death rate time series. To the best of our knowledge, this is the first time that Bi-GRU and Bi-Conv-LSTM models are used for prediction on COVID-19 new cases and new deaths time series. The evaluation of the methods is presented in the form of graphs and Friedman statistical test. The results show that the bidirectional models have lower errors than other models. Several error evaluation metrics are presented to compare all models, and finally, the superiority of bidirectional methods is determined. This research could be useful for organizations working against COVID-19 and determining their long-term plans.

## 1. Introduction

Serious Intense Respiratory Disorder Coronavirus 2 (SARS-COV-2) is a novel zoonotic microorganism (1). It is liable for Coronavirus Disease 2019 (COVID-19) (2, 3). The World Health Organization (WHO) and the worldwide countries affirmed the Covid-19 to be very infectious (4, 5). The spread rate of Covid-19 has increased day by day in numerous nations, particularly in United States (6), Spain (7), Italy (8), Germany (9), United Kingdom (10), France (11), and Iran (12). Estimating the prevalence of coronavirus is useful for controlling this pandemic. Each day more than 800,000 persons are infected by COVID-19 worldwide (13). The foremost challenging aspect of its spread is that individuals may be infected without having any symptoms explicitly for several days (14-16).

Machine Learning (ML) has demonstrated itself as a specific research field in recent decade by solving numerous exceptionally complex and advanced real-world problems (17, 18). In this research, the number of new cases and new deaths are predicted using deep learning which is a subfield of ML. There are existing literature which have tried to predict mortality each day. In this article, the prediction of mortality rate and new cases are performed every day, every three and seven days using deep learning models such as Long Short Term Memory (LSTM), Bidirectional-LSTM (Bi-LSTM), Convolutional-LSTM (Conv-LSTM), Bidirectional-Conv-LSTM (Bi-Conv-LSTM) and Gated Recurrent Unit (GRU), and Bidirectional-GRU (Bi-GRU).

The motivation of this research is preforming in-depth comparison of LSTM, Conv-LSTM, GRU with their bidirectional extensions. Moreover, based on the existing literature, it seems that Bi-GRU and Bi-Conv-LSTM have not been used before as predictors of COVID-19 time series data. During our experiments, we rely on Friedman test to compare the six deep learning methods statistically. Similar to the existing literature, we perform every day forecasting. Unlike the previous works, we also perform prediction every three and seven days which require one-third and one-seventh of every day prediction computational complexity. Investigation of prediction every three and seven days is done to determine whether it is possible to reduce computational complexity and still achieve reasonable performance. Computational complexity reduction matters in any application involving real-time forecasting of time series. The rest of the paper is structured as follows: Section 2 contains related research in this field, Section 3 reviews the background knowledge briefly, dataset description is provided in section 4, Section 5 is devoted to proposed method, Section 6 gives the experimental results, Section 7 presents discussion and Section 8 renders the conclusion and future works.

## 2. Related works

In this section, we briefly review the existing literatures that have similar scope with this paper. The differences between the reviewed works and our approach will be highlighted as well. Pinter et al. (19) predicted the number of infected people and the mortality rate by employing a hybrid ML approach. Their hybrid method consisted of a Multi-layered Perceptron (MLP) and Imperialist Competitive Calculation (MLP-ICA). The MLP was used as the predictor and ICA (an evolutionary optimization method) was used as the optimizer. The hybrid method was trained on Hungary dataset (20). The trained model was compared against adaptive network-based fuzzy inference system (ANFIS). The prediction horizon was chosen to be nine days.

Burke et al. (6) illustrated the ML model capability to determine the number of persons influenced by COVID-19 and the number of deceased cases. Linear Regression (LR), Least Absolute Shrinkage and Selection Operator (Lasso), Support Vector Machine (SVM), and Exponential Smoothing (ES) were utilized in their study (6). They showed that their method had the best performance among other similar methods.

Dowd et al. (21) investigated the effect of population age on the mortality rate of COVID-19 patients by utilizing numerical modeling. They reported that the infection is more life threatening to the older ages. Thus the approaches like social distancing and isolation can offer assistance to slow down and stop the spread of the virus.

Arun and Iyer (22) examined the prevalence of COVID-19 infection and anticipated the scale of the pandemic and mortality rate. They utilized ML and numerical modelling methods such as Polynomial Regression, Bayesian Edge and LSTM.

A study conducted by Zeroual et al. (23) proposed a deep learning system for prediction of COVID-19 time series. The main purpose of this study was to investigate deep learning methods for the number of deaths with limited information. The deep models can predict COVID-19 time series up to a specific horizon based on given time-variant inputs. The results showed that the Variational Auto Encoder (VAE) model outperformed other models.

Babaei et al. (24) analyzed the impact of health-protective measures such as quarantine, wearing masks, and social distancing using a susceptible-exposed-infectious-recovered (SEIR) type model on a hypothetical population. To further improve the model, the environmental noise (present in the data) has been taken into account using Brownian motion process. In addition, the stability analysis of the proposed model has been discussed. The authors reported that health strategies play a major role to contain the virus threat.

A mathematical model about the spread of COVID-19 was proposed in (25). The unique solvability of the proposed model was also proved. Additionally, the reproduction number of the proposed model was discussed. To survey the behavior of the considered model, some numerical simulations were conducted. Another research on the spread of COVID-19 has been conducted by Babaei et al. (26). The authors introduced a stochastic model considering several disease compartments related to different age groups. Their model was based on observing safety protocols, such as using mask and putting people into quarantine. The numerical results showed the effectiveness of safety protocols on COVID-19 containment.

Danane et al. (27) investigated the dynamics of COVID-19 stochastic model with isolation strategy. The authors relied on a SIQR (28) model and made it stochastic to take into account the uncertainty of infection progress. To this end, in all compartments of the proposed model, the white noise and the Levy jump perturbations were added. The existence and uniqueness of a global positive solution were proven and the stochastic dynamic properties of the solution around the deterministic model were investigated. The theoretical results were verified by some numerical simulations. While the authors relied on COVID-19 Morocco cases (29) to estimate the infection and the recovery rates of their simulations, we use Iran and Australia data (30) in our experiments. Another difference between the work

(27) and ours is the modeling approach. Danane et al. (27) used a stochastic version of SIQR to simulate dynamic of the virus while we rely on deep learning to carry out our predictions.

Singh et al. (31) analyzed the evolution of COVID-19 spread in an assumed population by employing a fractional-order dynamical system. They proposed a stable computational method to solve the dynamical system numerically. The computational method is based on the discretization of the domain and the short memory principle. The implemented approach divided the population to five subgroups such as susceptible people, exposed people, infected people, etc. and analyzed how these subgroups behave over time. Gao et al. conducted another study to describe COVID-19 spread behavior based on fractional calculus (31, 32). They utilized fractional natural decomposition (FNDM) to understand the dynamical structure of COVID-19. The methods in (32) have analyzed COVID-19 spread behavior well but working with fractional-order systems involves more complex computation compared to neural networks. Gao et al. (33) have also employed fractional calculus to clearly describe the reported and unreported cases of COVID-19. To this end, a time-fractional model was parameterized using reported cases of the virus. The model solution was found by the q-homotopy analysis transform method. The number of unreported cases of the virus was then identified. They were able to predict the exponential growth of the virus using their model. The three methods reviewed above ((31-33)) are based on fractional calculus and have well-established mathematical foundations; however, they are not easy to grasp and implement for the general readers. Our method on the other hand is based on neural networks which is more intuitive and easier to work with.

Boudaoui et al. (34) have relied on Caputo–Fabrizio fractional derivative to extend the transmission model of COVID-19 proposed by Tang et al. (35). The existence and uniqueness of solution for the extended model has been discussed and the solution has been obtained using a numerical approach. Based on the conducted simulations using the model, the authors reported that the infective population peak decreases as the contact rate is decreased and isolation/hospitalization of infected individuals is increased. Despite presenting interesting results, the method proposed by Boudaoui et al. is based on a fixed mathematical model which may bias the simulation results. On the contrary, we rely on the data collected from the population in a dynamic manner and use them during training and prediction of our neural network-based model. Therefore, our model is able to adapt to the changing dynamics of the population on the fly which reduces the bias in its prediction.

Zamir et al. (36) took a Non Pharmaceutical Intervention (NPI) approach to reduce the outbreak of COVID-19. To this end, the population concerned with the disease was divided to six compartments based on which a mathematical model consisting of coupled differential equations was proposed. Analyzing the model, they were able to determine NPIs critical to the virus containment. The important NPIs were isolation, sanitizers, infection side effects treatment, and wearing face mask. While Zamir et al. focused on devising strategies to flatten COVID-19 infection curve, we focus on forecasting the mortality and spread of the virus.

Facing COVID-19 without having an effective vaccine, many governments panicked and adopted lock down strategy to prevent the virus spread. However, such a strategy hurts the global economy. Sahoo et al. (37) investigated the possibility of containing the virus without lock down. To this end, mathematical models based on partial differential equations were considered to inspect the effect of proper quarantine with no lock down on the virus spread. The authors reported that social distancing and proper quarantine of citizens prior to entering their native countries or native states are the best preventive measures in the absence of vaccine. While Sahoo et al. tried to determine general measures to prevent the virus spread; we aim to predict the trend of the virus spread and mortality.

Gao et al. (38) investigated the numerical distributions of COVID-19 according to time. To this end, the authors found the optimal values for the mathematical model Bats-Hosts-Reservoir-People coronavirus (BHRPC) of the virus transfer from the reservoir to people. The Variational Iteration Method (VIM) was employed for numerical investigation of BHRPC model. To reach realistic results, the model parameters were chosen according to the values reported by experts in Wuhan area of China. The authors reported that presence of susceptible people in the population accelerates the virus spread. While Gao et al. (38) focused on the virus transfer from the reservoir to people, we focus on prediction on mortality rate and the spread of the virus based on observed data.

## 3. Deep Learning and its variations

Deep learning (DL) is a machine learning algorithm which is based on artificial neural networks (ANNs). This research introduces a DL system for the prediction of the COVID-19 time series. The following is an introduction to some of the DL methods used to predict time series namely LSTM, Bi-LSTM, Conv-LSTM and GRU.

LSTM is a special type of Recurrent Neural Network (RNN) which relies on its repeating module called cell to remember sequence of information. Each cell contains three gates namely input, output, and forget gates. The forget gate decides how much information of the cell state must be thrown away. The input gate specifies the new information that must be stored in the cell state. The output gate decides the parts of the cell state that must be sent to the cell output.

A Bi-LSTM network is an extension of traditional LSTM which trains two LSTMs. One of the LSTMs is trained on the input sequence. The other LSTM is trained on the input sequence but in reversed order. Bi-LSTM can achieve faster learning compared to traditional LSTM.

Traditional LSTM has been designed to work with one-dimensional data so it cannot cope with multi-dimensional data such as images. Conv-LSTM replaces the associated gate layers of the LSTM with convolutional layers to address this issue. Conv-LSTM can encode Spatio-temporal data in its memory cell (39). Subsequently, by supplanting the convolution operators with an LSTM memory cell, the Conv-LSTM can know which data should be 'remembered' or 'forgotten' from the past cell state.

GRU (40) is a special version of RNN. GRU is similar to LSTM but instead of three, the number of gates in GRU is two: upgrade and reset gates. The upgrade gate determines how the past information should be passed along to the future. The reset gate determines how much of the past information must be discarded (41).

## 4. Dataset Description

This research aims to predict COVID-19 prevalence in the future, focusing on the new cases and the new deaths rate. The dataset used in this research contains the statistical reports of COVID-19 cases and the mortality rate of different countries. It has been obtained from the WHO website (42). The dataset includes eight different columns such as "Date Reported", "Country Code", "Country", "WHO Region", "Cumulative Cases", and "Cumulative Deaths". In this research, "New Cases", "Cumulative Cases", "New Deaths", and "Cumulative Deaths" columns are used as time series to forecast future rate of new cases and deaths in Australia and Iran. The rest of the features are presented in Table 1. In the presented study, data from two countries Australia and Iran are used.

**Table 1**: Data Description

| Date Reported | Country Code | Country | WHO Region |
|---|---|---|---|
| 1/25/2020 – 8/19/2020 | AU | Australia | Western Pacific Regional Office (WPRO) |
| 1/3/2020 – 10/6/2020 | IR | Iran | Eastern Mediterranean Regional Office (EMRO) |

## 5. Proposed Method

In this research, a DL-based approach was used to forecast the rate of new cases and new deaths every one, three and seven days. We experimented with six neural network models as our predictor. Each model consists of an input layer, an output layer and three hidden layers. The first three models were LSTM, Conv-LSTM, and GRU. The next three models were the bidirectional version of the first three ones i.e. Bi-LSTM, Bi-Conv-LSTM and Bi-GRU. The number of neurons in the hidden layers was 50. In all layers, the Rectified Linear Unit (ReLU) was used as the activation function. The training was performed with respect to MSE loss function using Adam optimiser. The hyper-parameters of Adam were set to $\beta_1$=0.9 and $\beta_2 = 0.999$. The learning rate was set to 0.001. The model was trained for 200 epochs. In Table 2, additional details of the implemented models are shown.

For the training data, the time series of Australia and Iran have been chosen from WHO website's database which reports new cases and new deaths rates. Approximately 70% of the data were used for training and the rest were kept for testing. About 20% of the training data were used for validation.

During the training for the first time, the time series were fed to the model based on which the model predicted the next day. The model training was repeated for the second time such that its output predicted the next three days. Finally, the model was trained for the third time to achieve predictions for the next seven days. As the forcasting horizon increases from one to three and to seven, the error rate of the model icreases which makes sense since forecasting for a longer horizon is harder than forecasting for a shorter horizon. The training process was implemented for both the time series of new cases and new deaths. Figure 1 illustrates the high leve steps of the proposed method.

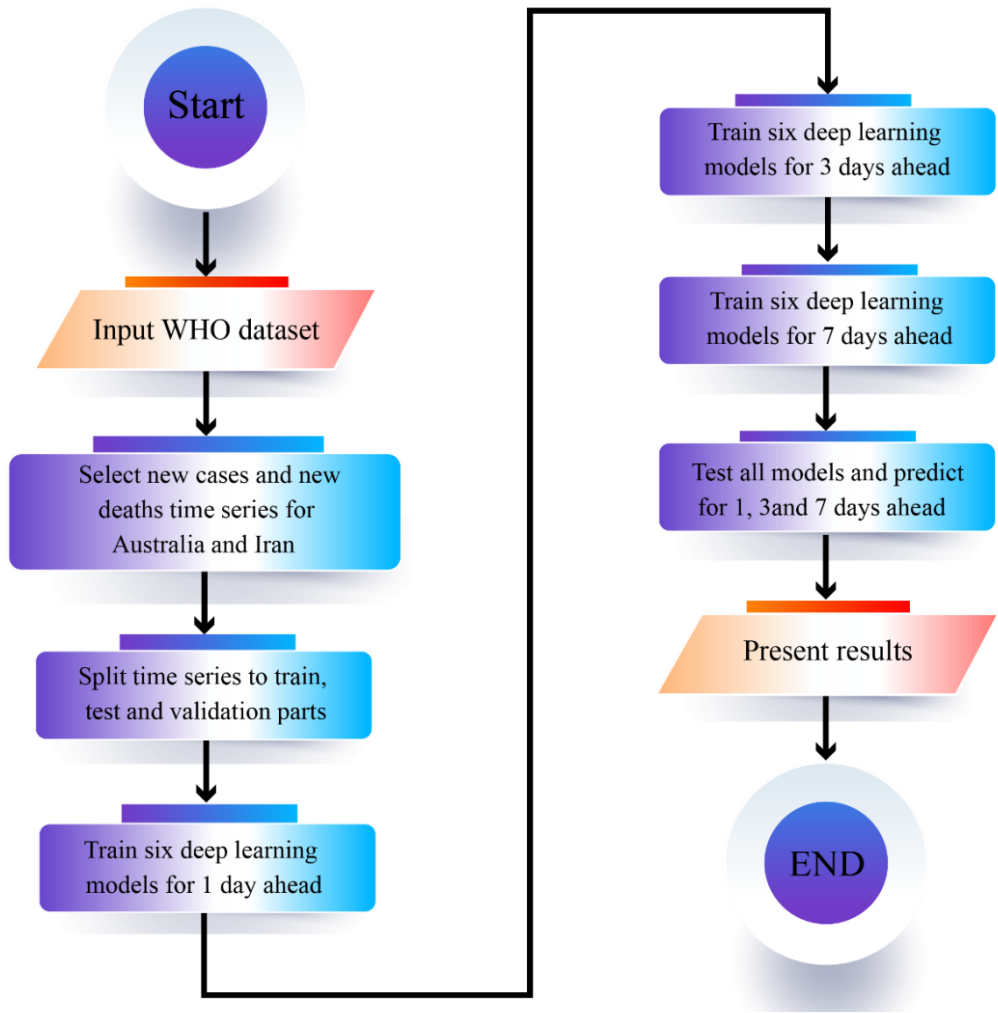

**Figure 1**: The proposed method high level steps

**Table 2**: Additional implementation details of the six models

| Model | Number of Hidden Layers | Number of Units | Number of Convolution Filters | Size of Convolution Kernels |
|---|---|---|---|---|
| LSTM | 3 | 50 | - | - |
| Bi-LSTM | 3 | 50 | - | - |
| Conv-LSTM | 3 | - | 64 | 1×2 |
| Bi-Conv-LSTM | 3 | - | 64 | 1×2 |
| GRU | 3 | 50 | - | - |
| Bi-GRU | 3 | 50 | - | - |

## 6. Experimental Results and Analysis

In this section, the experimental results for LSTM, Conv-LSTM and GRU as well as their bidirectional counterparts are reported. To the best of our knowledge, we are the first to use

Bi-Conv-LSTM and Bi-GRU for prediction of COVID-19 new cases and deaths based on time series data.

To have a fair comparison, we tried to implement all methods with relatively similar conditions. The prediction error was calculated based on critieria (14) such as Mean Squared Log Error (MSLE), Mean Absolute Percentage Error (MAPE), Root Mean Squared Log Error (RMSLE), and Explained Variance (EV). These evaluation criteria are computed as below(41):

$$MSLE = \frac{1}{n}\sum_{i=1}^{n}(\log(y_i) - \log(\hat{y}_i))^2, \qquad (1)$$

$$MAPE = \frac{100}{n}\sum_{i=1}^{n}\left|\frac{y_i - \hat{y}_i}{y_i}\right|, \qquad (2)$$

$$RMSLE = \sqrt{\frac{1}{n}\sum_{i=1}^{n}(\log(y_i) - \log(\hat{y}_i))^2}, \qquad (3)$$

$$EV = 1 - \frac{\text{Var}(\hat{y}_i - y_i)}{\text{Var}(y_i)}, \qquad (4)$$

where $y_i$ is the actual values, $\hat{y}_i$ is the corresponding estimated values, and n is the number of samples.

*6.1. Forecasting performance*

For each of the mentioned methods, the error of 1, 3, and 7-day ahead predictions for new cases/deaths in a 100-day period were calculated in Australia and Iran. To this end, the predicted values were compared with the actual values, and the error rate was calculated based on evaluation criteria (Equations 1-4). The results of calculating the errors in the 100-day period for each of Australia's models are given in Figure 2 and Figure 3. As it is apperant from MAPE values in Figure 2, Bi-GRU and LSTM have the best performance in the 1-day perdiction, Conv-LSTM is the best method in the 3-day prediction, and Bi-Conv-LSTM has the best performance in the 7-day prediction. All of the evaluated methods in Figure 2 have approximately similar explained variance. Figure 3 illustrates the evaluation results for new deaths prediction for the 100-day period in Australia. An interesting observation in Figure 3 is how LSTM significantly outperforms GRU in the 7-day ahead prediciton. The reason lies in the fact that GRU has a simpler structure (less parameters) consisting of only two gates. However, the more complex structure of LSTM seems to prevail sometimes as it is the case in the 7-day prediciton of new deaths in Figure 3.

The evaluation results of the models for prediction of Iran new cases and new deaths are presented in Figures 4 and 5, respectively. The main observation based on MAPE criterion in these figures is that most of the time LSTM and its variations outperform GRU especially for the longer horizons (3 and 7-day) scenarios.

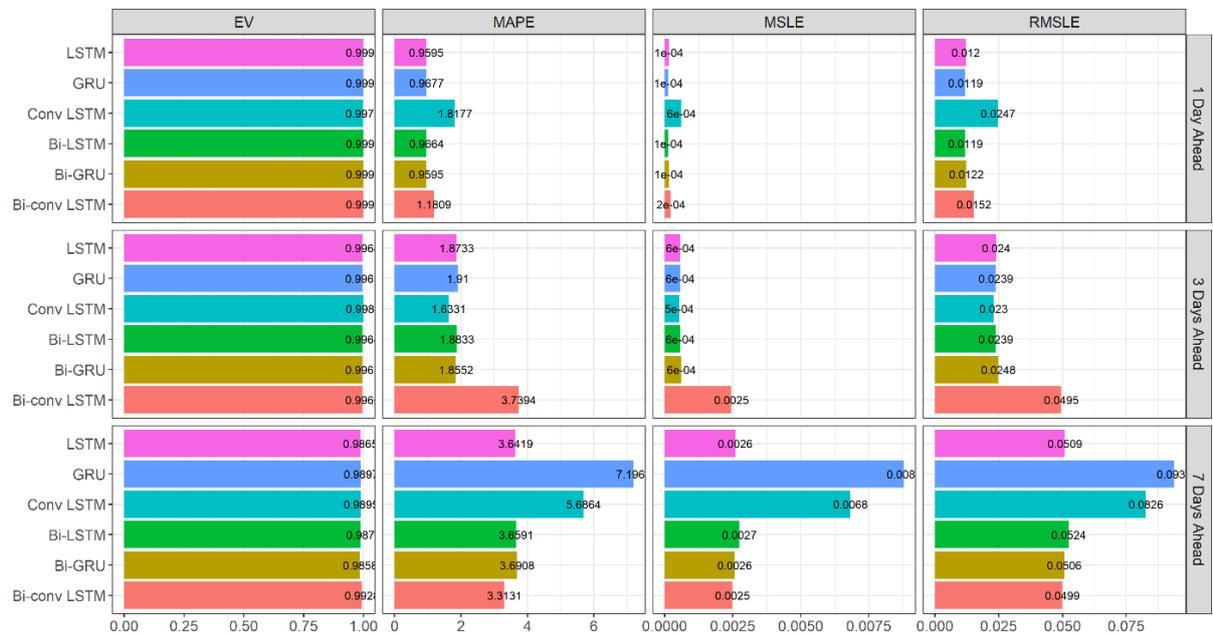

**Figure 2**: Evaluation metrics for new cases forecasting in Australia

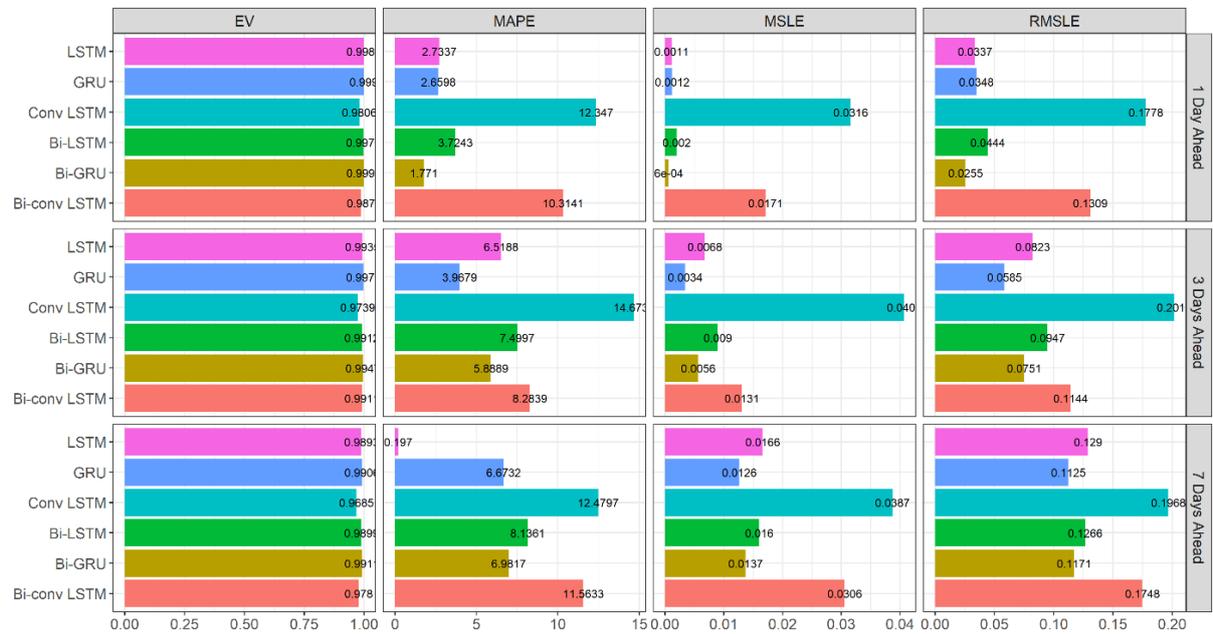

**Figure 3**: Evaluation metrics for new deaths forecasting in Australia

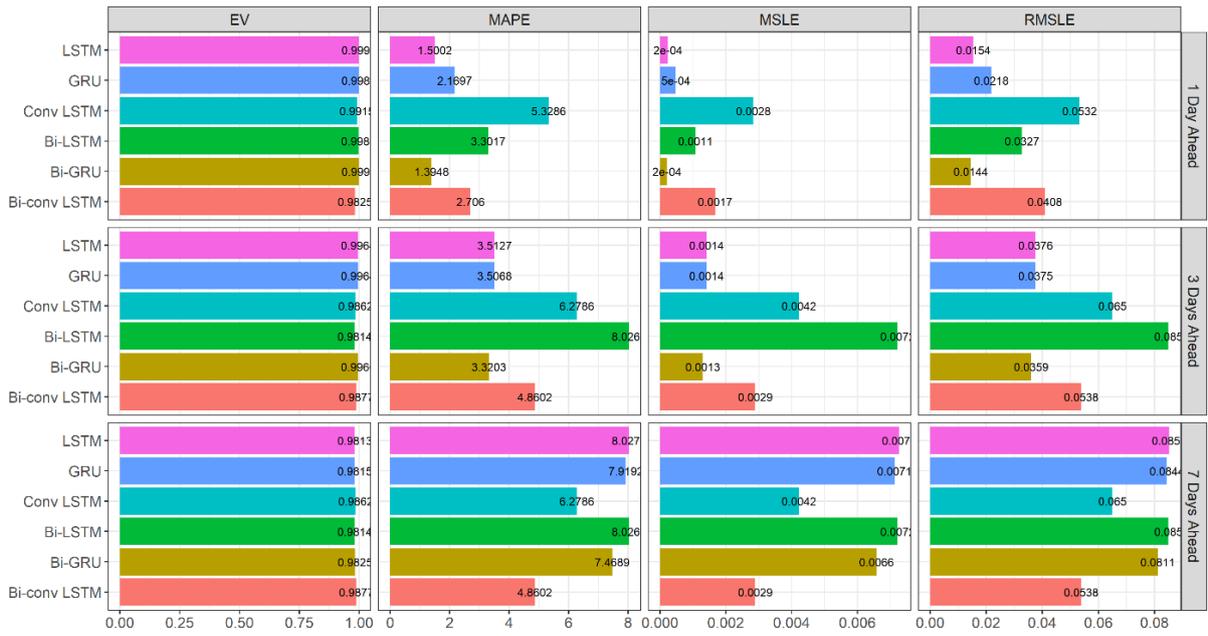

**Figure 4**: Evaluation metrics for new cases forecasting in Iran

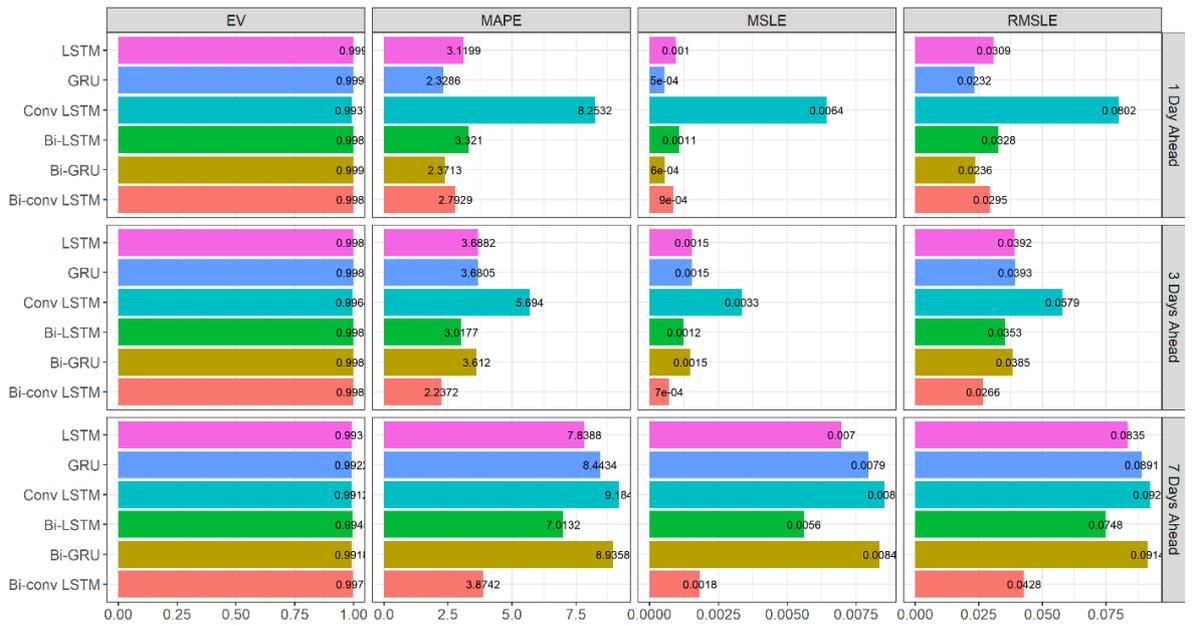

**Figure 5**: Evaluation metrics for new deaths forecasting in Iran

The predictions of new cases and new deaths in Australia and Iran for 1, 3, and 7-day ahead are compared with actual data in Figures 6-9. The prevalent pattern observed in these figures is that longer prediction horizon often leads to larger prediction errors. Of course such pattern is violated in Figure 8.a where Bi-Conv-LSTM has deviated from the actual data badly.

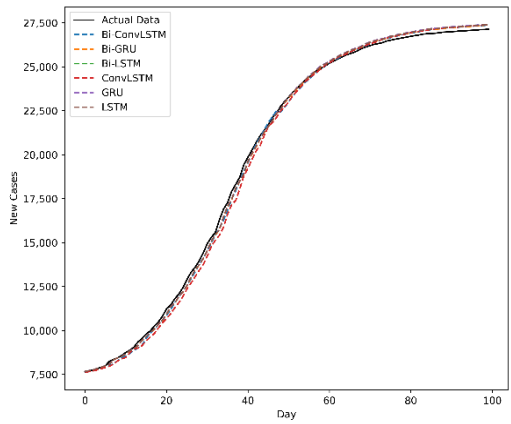
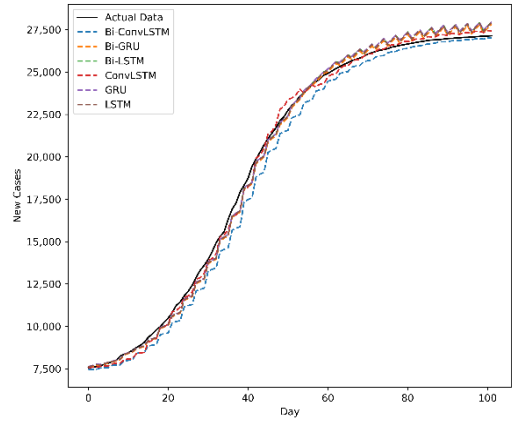

a

b

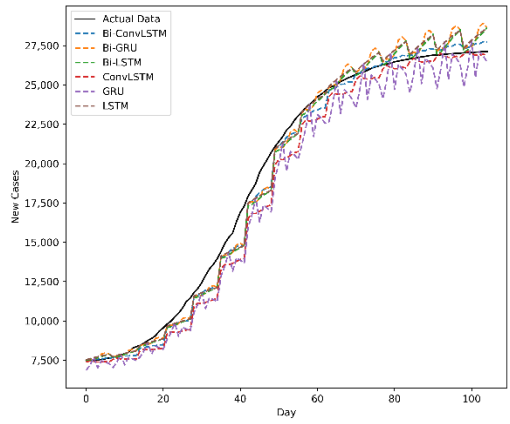

c

**Figure 6:** New cases forecasting a) every day, b) every 3 days and c) every 7 days in Australia

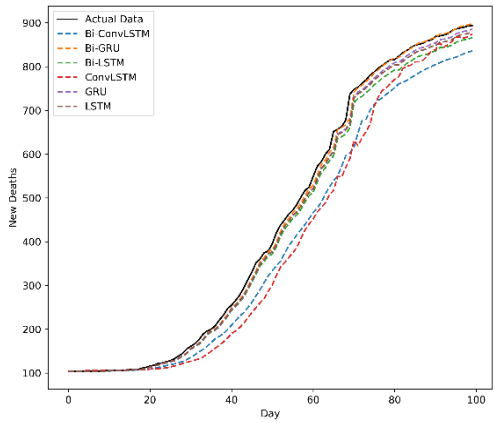
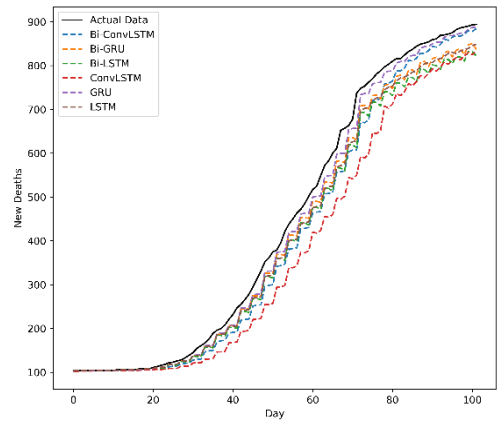

a

b

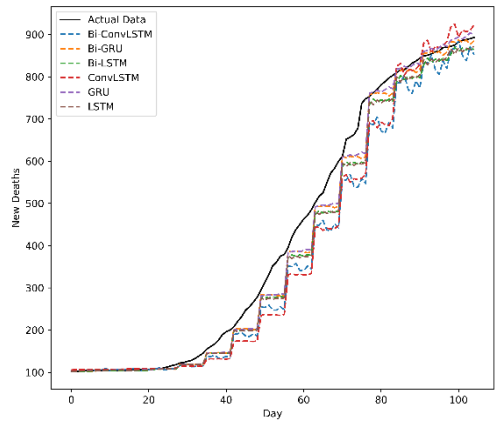

c

**Figure 7:** New deaths forecasting a) every day, b) every 3 days and c) every 7 days in Australia

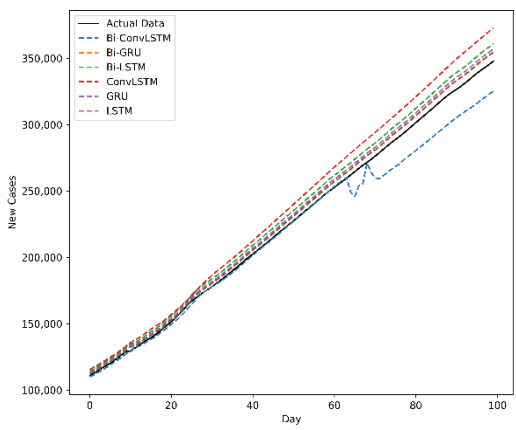

a

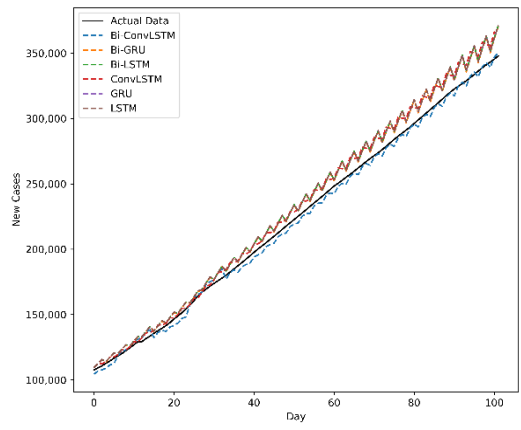

b

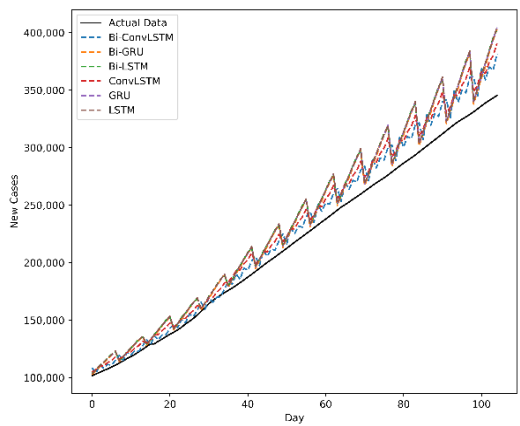

c

**Figure 8:** New cases forecasting a) every day, b) every 3 days and c) every 7 days in Iran

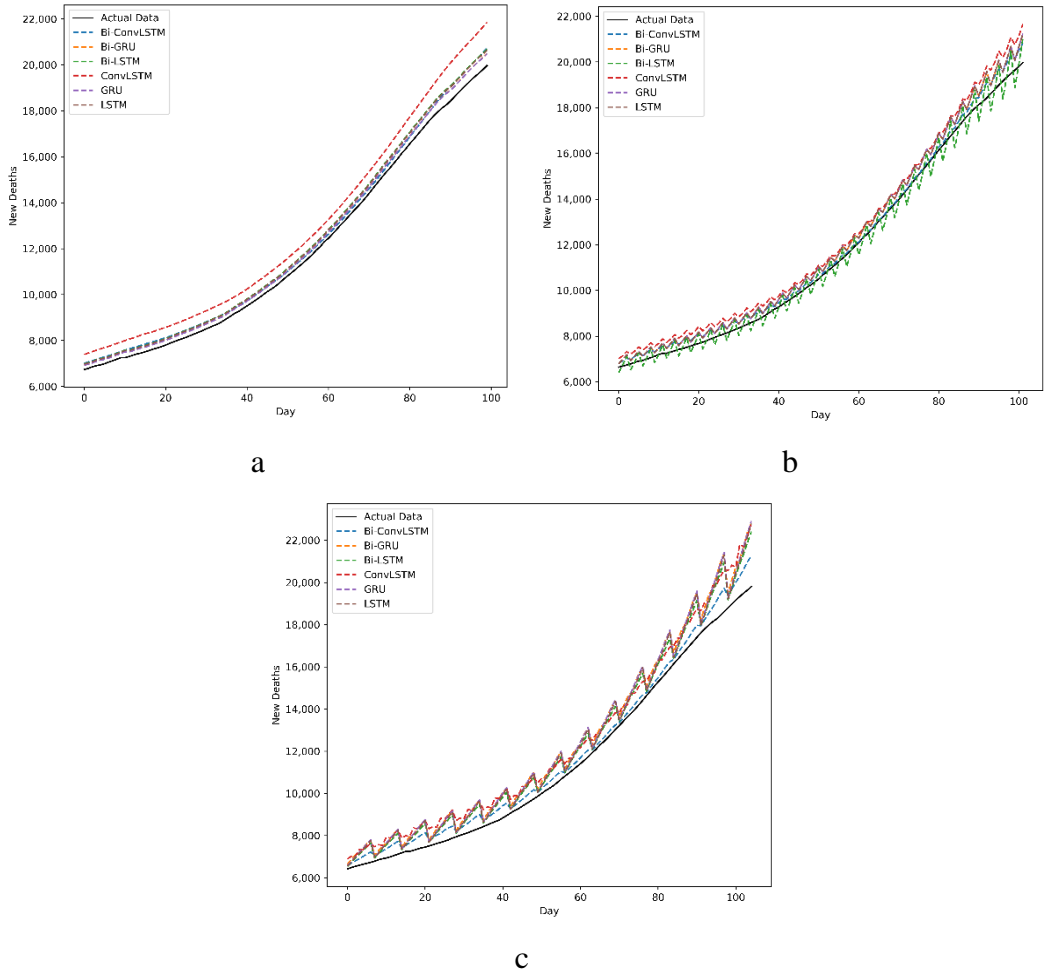

**Figure 9**: New deaths forecasting a) every day, b) every 3 days and c) every 7 days in Iran

For better comparison of the six methods performance, histograms of their absolute error are also presented in Figures 10-13. The horizontal axes of the histograms represent the absolute error which is the difference between the models predictions and the actual data. Recall that predictions are performed for 100 days. The height of each histogram bin shows the number of predictions which absolute error falls within the interval dictated by that bin. For better clarity, in Figures 10-13, the histograms of different methods are drawn with different colours and line widths. At first glance, the maximum absolute error observed in Figures 10-13 seems to be too high. However, it should be noted that the number of new cases/deaths in each day are cumulative values. In other words, the actual new cases/deaths reported for i-th day is the total number of new cases/deaths reported from first day until i-th day across the whole country (such as Iran). Considering that data are reported as cumulative values, it comes as no surprise that they are usually large values. The absolute error of prediction is directly influenced by the magnitude of the actual data. For example in one day prediction of new cases for Iran (Figure 12.a), GRU has predicted 354000 while true value was 345000. The absolute error in this case is $|345000 - 354000| = 9000$ which is 2.6% of the true value. Therefore, at first 9000 seems to be a large error but it is indeed tolerable compared the magnitude of the true value

(345000). For the seven days ahead prediction (Figure 12.c), the same method has predicted 404000 while the true value was 345000. The absolute error is |345000 − 404000| = 59000 which is 17.10% of the true value. Obviously, the absolute error of the seven days ahead prediction is higher than that of one day ahead prediction. However, 59000 is still a reasonable value compared to the true value.

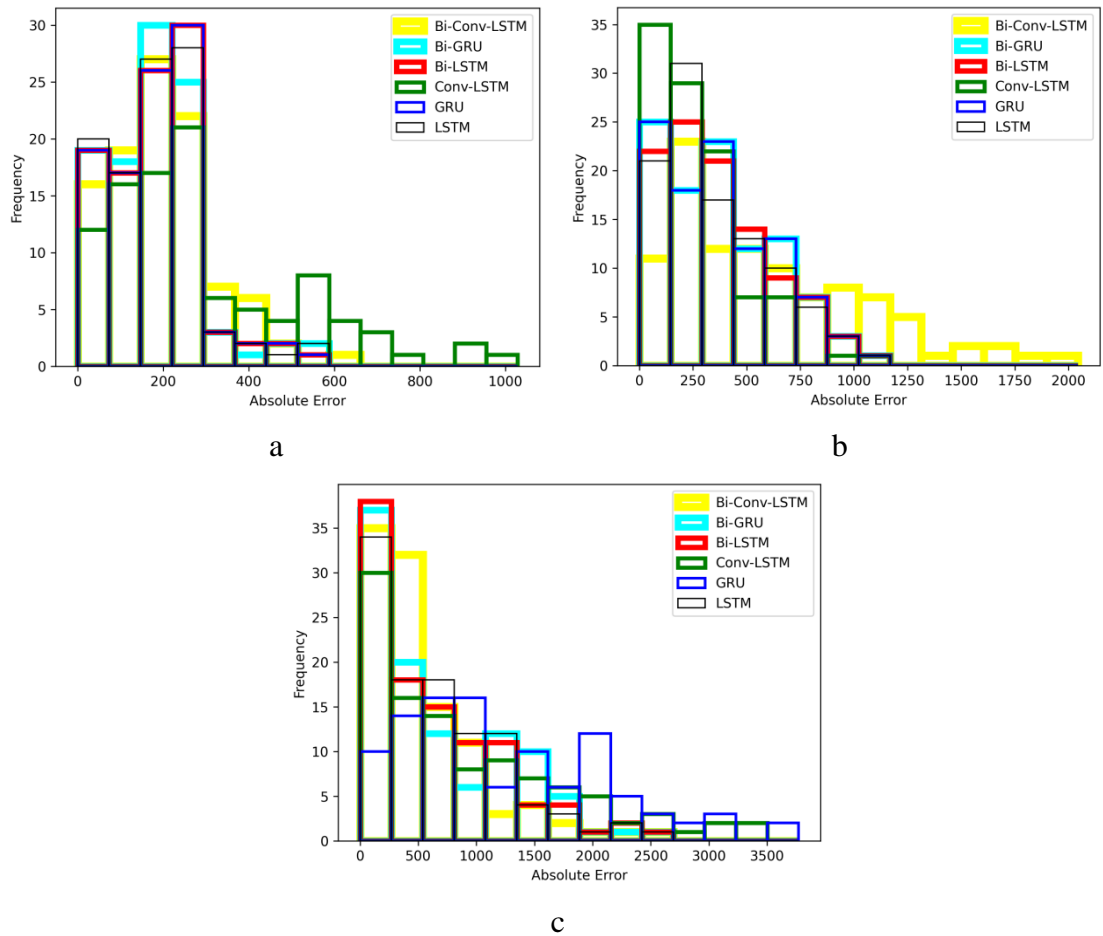

**Figure 10**: Absolute error histogram of forecasting new cases in Australia a) every day, b) every 3 days and c) every 7 days

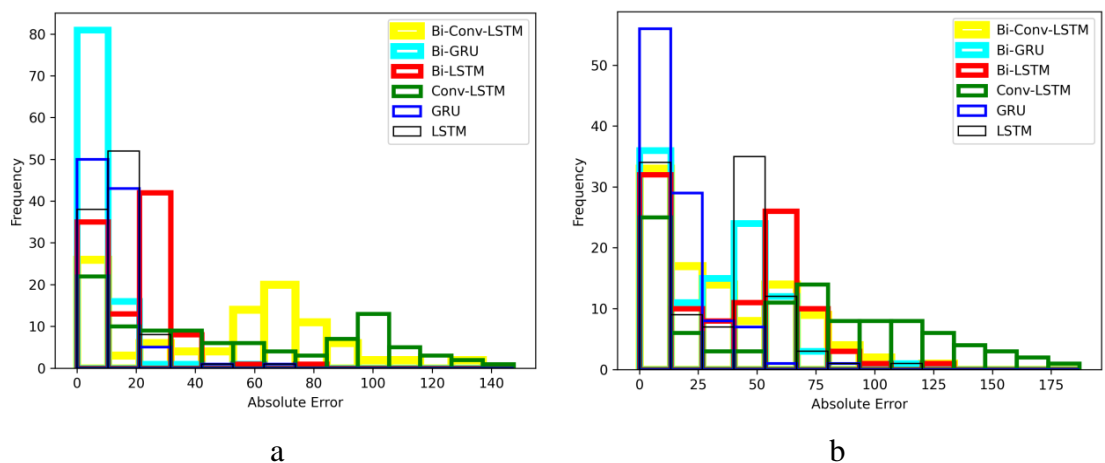

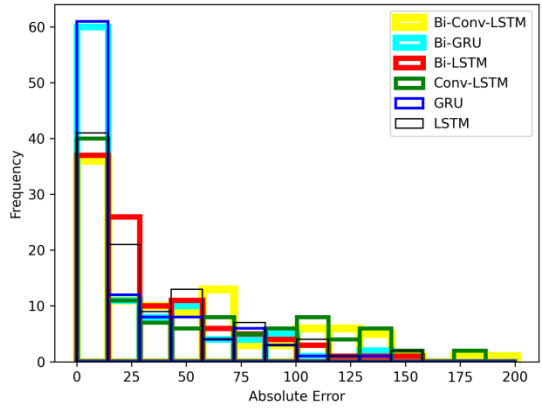

c

**Figure 11**: Absolute error histogram of forecasting new deaths in Australia a) every day, b) every 3 days and c) every 7 days

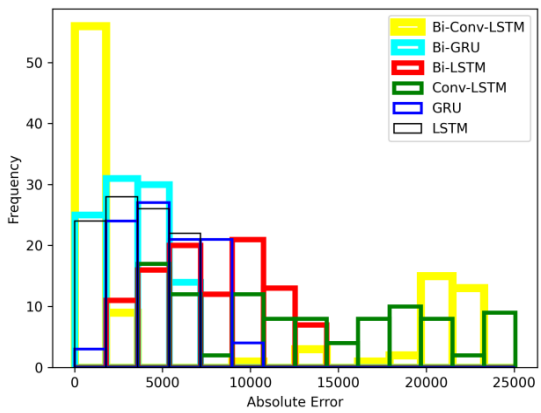

a

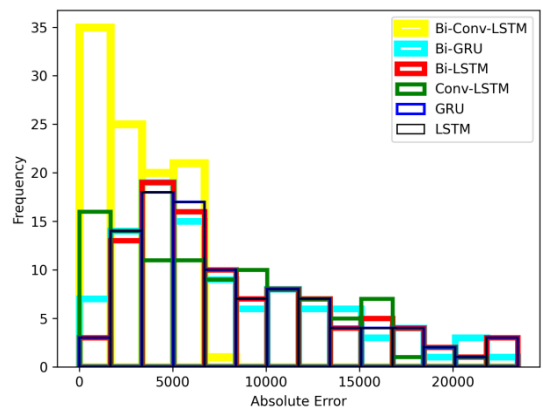

b

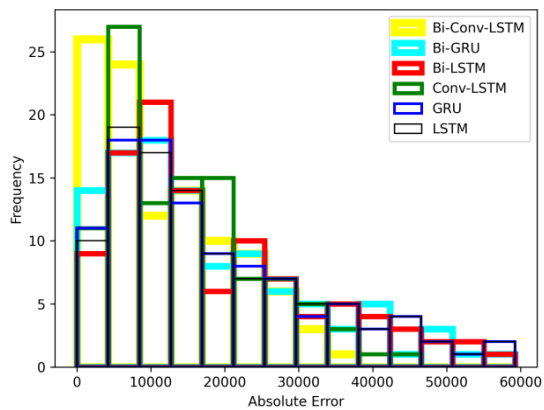

c

**Figure 12**: Absolute error histogram of forecasting new cases in Iran a) every day, b) every 3 days and c) every 7 days

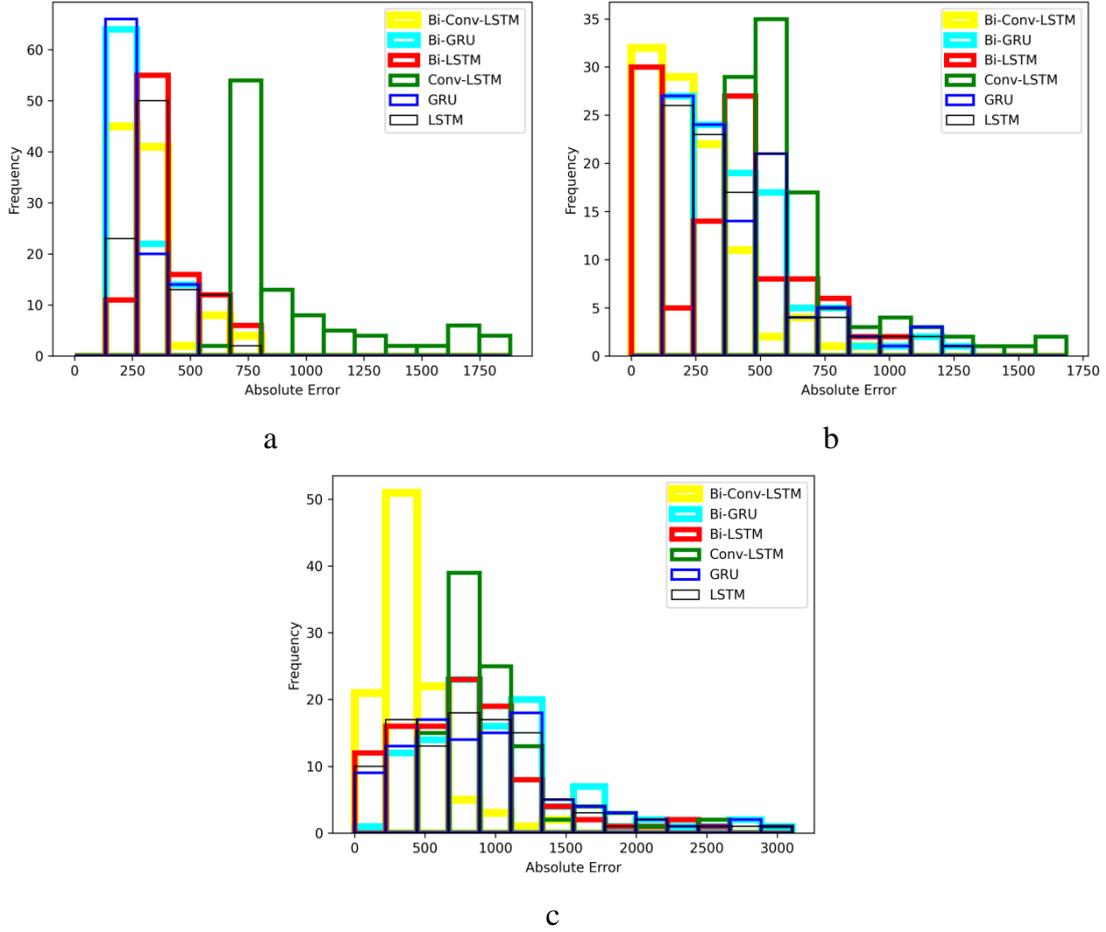

**Figure 13**: Absolute error histogram of forecasting new deaths in Iran a) every day, b) every 3 days and c) every 7 days

## 6.2. Statistical Analysis

Friedman (43, 44) proposed a non-parametric statistical test known as the Friedman test which is widely used by researchers to analyse their methods (45). In this subsection, Friedman test was used to compare the algorithms. To this end, the average of error evaluation criteria (EV, MAPE, MSLE, RMSLE) in Figures 2-5 were computed and listed in Table 3 based on which the algorithms were ranked as shown in Table 4. The methods with lower ranks are better than the ones with higher ranks.

To carry out Friedman test, the rankings from Table 4 are required. Suppose the rank of the j-th classifier on the i-th dataset is denoted by $r_i^j$ so the average rank of the algorithms can be computed by $R_j = \frac{1}{N}\Sigma r_i^j$. The Friedman test is then computed by the following formula:

$$X_F^2 = \frac{12N}{k(k+1)} \left[ \sum_j R_j^2 - \frac{k(k+1)^2}{4} \right] \quad (5)$$

where k is the number of algorithms and N is the number of datasets. Motivated by Friedman test, Iman and Davenport (46) proposed another statistical test as follows:

$$F_f = \frac{(N-1)X_F^2}{N(k-1)-X_F^2}, \qquad (6)$$

which has F-distribution with $((k - 1),(k - 1)(N - 1))$ degrees of freedom. According to the results of Table 4, $\chi^2_F$ and $F_f$ are computed as follows:

$$X_F^2 = \frac{12 \times 12}{6(6+1)} \left[ (3)^2 + (3.25)^2 + (4.83)^2 + (4.08)^2 + (2.33)^2 + (3.42)^2 - \frac{6(6+1)^2}{4} \right] = 10.84$$

$$F_f = \frac{(12-1) \times 10.84}{12(6-1) - 10.84} = 2.43$$

According to six algorithms and 12 datasets (New cases 1-day AU, …), $F_f$ is governed by the F-distribution with $((k - 1), (k - 1)(N - 1)) = (5,55)$ degree of freedom. The critical value of F(5,55) is 2.38 for significance level α = 0.05. As it is clear in Table 4, Bi-GRU algorithm has the best average rank among all the algorithms followed by LSTM, GRU, Bi-Conv-LSTM, Bi-LSTM, and Conv-LSTM.

**Table 3:** Average of error evaluation metrics

| Dataset | LSTM | GRU | Conv-LSTM | Bi-LSTM | Bi-GRU | Bi-Conv-LSTM |
|---|---|---|---|---|---|---|
| New Cases 1-day AU | 0.49265 | 0.494675 | 0.71 | 0.49435 | 0.4927 | 0.548825 |
| New Cases 3-day AU | 0.723475 | 0.732625 | 0.66365 | 0.72595 | 0.71915 | 1.19685 |
| New Cases 7-day AU | 1.170475 | 2.074175 | 1.691325 | 1.18285 | 1.1824 | 1.0894 |
| New Deaths 1-day AU | 0.941625 | 0.9237 | 3.38425 | 1.191925 | 0.699025 | 2.862275 |
| New Deaths 3-day AU | 1.900225 | 1.2567 | 3.971975 | 2.14865 | 1.7409 | 2.3506 |
| New Deaths 7-day AU | 0.33295 | 1.947175 | 3.420925 | 2.317075 | 2.025875 | 3.186675 |
| New Cases 1-day IR | 0.6287 | 0.7975 | 1.594025 | 1.083375 | 0.6021 | 0.93275 |
| New Cases 3-day IR | 1.136925 | 1.135425 | 1.8335 | 2.269875 | 1.088375 | 1.476 |
| New Cases 7-day IR | 2.275075 | 2.24805 | 1.8335 | 2.269875 | 2.124775 | 1.476 |
| New Deaths 1-day IR | 1.0377 | 0.852075 | 2.3332 | 1.088225 | 0.848625 | 0.955325 |
| New Deaths 3-day IR | 1.181725 | 1.179825 | 1.6878 | 1.01305 | 1.1625 | 0.815625 |
| New Deaths 7-day IR | 2.230575 | 2.3831 | 2.56875 | 2.0219 | 2.50665 | 1.22895 |

**Table 4:** Rank of the algorithms on datasets.

| Dataset | LSTM | GRU | Conv-LSTM | Bi-LSTM | Bi-GRU | Bi-Conv-LSTM |
|---|---|---|---|---|---|---|
| New Cases 1-day AU | 1 | 4 | 6 | 3 | 2 | 5 |
| New Cases 3-day AU | 3 | 5 | 1 | 4 | 2 | 6 |
| New Cases 7-day AU | 2 | 6 | 3 | 5 | 4 | 1 |
| New Deaths 1-day AU | 3 | 2 | 6 | 4 | 1 | 5 |
| New Deaths 3-day AU | 3 | 1 | 6 | 4 | 2 | 5 |
| New Deaths 7-day AU | 1 | 2 | 6 | 4 | 3 | 5 |
| New Cases 1-day IR | 2 | 3 | 5 | 5 | 1 | 4 |
| New Cases 3-day IR | 3 | 2 | 1 | 6 | 1 | 4 |
| New Cases 7-day IR | 6 | 4 | 6 | 5 | 3 | 1 |
| New Deaths 1-day IR | 4 | 2 | 6 | 5 | 1 | 3 |
| New Deaths 3-day IR | 5 | 4 | 6 | 2 | 3 | 1 |
| New Deaths 7-day IR | 3 | 4 | 6 | 2 | 5 | 1 |
| Average Rank | 3 | 3.25 | 4.83 | 4.08 | **2.33** | 3.42 |

## 7. Discussion

Time series prediction is an important topic in finance, economics, and business. Recent advancement in computers' computational power, ML methods and new perspectives such as DL has led to the emergence of new algorithms for times series analysis and prediction. Some of these algorithms are LSTM, GRU, Conv-LSTM, Bi-LSTM, Bi-GRU and Bi-Conv-LSTM. Each algorithm has its advantages and disadvantages. Our investigation about the forecasting ability of these methods on COVID-19 time series led to the following contributions:

- Based on the literature review, it seems that Bi-GRU and Bi-Conv-LSTM models have never been used before for prediction on COVID-19 new cases and new deaths rate time series.

- No research was found which predicts new cases and new deaths every three or seven days. The motivation behind attempting to predict every n days (instead of every day) was to investigate whether it is possible to reduce computational complexity and still achieve reasonable performance. Such a scenario gains importance in any application involving real-time forecasting of time series. Whether the incurred error due to prediction every n days is acceptable or not, fully depends on the application requirements. Therefore, it is the designer who decides whether it is worth to sacrifice performance to gain better computation efficiency. In our experiments, inspection of RMSLE metric in figures 2-5 shows that predicting every three days approximately doubles the prediction error. The incurred error of predicting every seven days is more than four times of error when prediction is done every day.

- Comprehensive evaluation of LSTM, Conv-LSTM, GRU and their bidirectional extensions.

- Statistical comparison of the investigated methods using Friedman test.

Recall that in Figures 2-5, the error rate of the new cases and new deaths in Iran and Australia were determined by methods LSTM, GRU, Conv-LSTM, Bi-LSTM, Bi-GRU and Bi-Conv-LSTM. Overall, it was observed that in most of the conducted experiments (Figures 2-5) the bidirectional methods achieved better results than the other methods.

Based on data in Table 4, the key observations can be summarized as below:

- For prediction of new deaths in the next day in Australia and Iran, Bi-GRU had the best performance. For 3-day ahead prediction of new deaths in Australia, GRU was the best method while Bi-Conv-LSTM did the best prediction in Iran. Finally, in the 7-day ahead case, LSTM performed better than other methods on Australia data and Bi-Conv-LSTM outperformed other methods on Iran data.

- On the other hand, for 1-day ahead predictions of new cases in Australia, LSTM and Bi-GRU gained the best performance. For 3 and 7-day ahead predictions, Conv-LSTM and Bi-Conv-LSTM showed better performance, respectively. In Iran, Bi-GRU was better for 1 and 3-day ahead predictions and Bi-Conv-LSTM was better for 7-day ahead prediction.

The proposed method can provide the health crisis management centers with valuable forecasting based on the observed data. Having an estimate of what awaits us in the near future might help with the appropriate preparation to minimize the inevitable damage. The forecasting ability of the six models is due to their memorizing capability. The limitation of the proposed method is that the characteristics of the time series data might change as time passes. Therefore, to keep the models accurate, we are forced to incur the cost of training the models on the newly observed data.

**8. Conclusion**

In this research, six different models were compared for predicting the number of new cases and deaths in the next 100 days. The prediction was done for each day, every 3 days and every 7 days. The conducted experiments showed that most of the time, the bidirectional models outperform their non-bidirectional counterparts.

In the future, the plan is to use a combination of other machine learning and deep learning methods to achieve better results. In particular, experimenting with non-parametric models such as Gaussian Process (GP) to perform time series forecasting seems interesting since GP can provide uncertainty about its predictions. We might be able to determine the appropriate prediction horizon based on the uncertainty provided by GP.

**List of Abbreviations**

| | |
|---|---|
| ANFIS | Adaptive Network-based Fuzzy Inference System |
| ANN | Artificial Neural Network |
| AU | Australia |
| Bi-GRU | Bidirectional Gated Recurrent Unit |
| Bi-Conv-LSTM | Bidirectional Convolutional Long Short Term Memory |
| Bi-LSTM | Bidirectional Long Short-Term Memory |
| Conv-LSTM | Convolutional Long Short Term Memory |
| COVID-19 | Coronavirus Disease 2019 |
| DL | Deep Learning |
| DLSTM | Delayed Long Short-Term Memory |
| EMRO | Eastern Mediterranean Regional Office |
| ES | Exponential Smoothing |
| EV | Explained Variance |
| GRU | Gated Recurrent Unit |
| IR | Iran |
| Lasso | Least Absolute Shrinkage and Selection Operator |
| LR | Linear Regression |
| LSTM | Long Short-Term Memory |
| MAE | Mean Absolute Error |
| MAPE | Mean Absolute Percentage Error |
| MSE | Mean Square Error |
| MERS | Middle East Respiratory Syndrome |
| ML | Machine Learning |
| MLP-ICA | Multi-layered Perceptron-Imperialist Competitive Calculation |
| MSLE | Mean Squared Log Error |
| PRISMA | Preferred Reporting Items for Precise Surveys and Meta-Analyses |
| ReLU | Rectified Linear Unit |
| RMSE | Root Mean Square Error |
| RMSLE | Root Mean Squared Log Error |
| RNN | Repetitive Neural Network |
| SARS | Serious Intense Respiratory Disorder |
| SARS-COV | SARS coronavirus |
| SARS-COV-2 | Serious Intense Respiratory Disorder Coronavirus 2 |
| SVM | Support Vector Machine |
| VAE | Variational Auto Encoder |
| WHO | World Health Organization |
| WPRO | Western Pacific Regional Office |

**CRediT author statement:**

**Nooshin Ayoobi:** Conceptualization, Data curation, Formal analysis, Investigation, Methodology, Validation, Writing – original draft **Danial Sharifrazi:** Conceptualization, Data curation, Formal analysis, Investigation, Methodology, Validation, Writing – original draft **Roohallah Alizadehsani:** Formal analysis, Methodology, Writing - original draft **Afshin Shoeibi:** Formal analysis, Methodology, Writing - original draft **Juan M. Gorriz:** Validation, Conceptualization, Investigation **Hossein Moosaei:** Formal analysis, Methodology, Writing - original draft **Abbas Khosravi:** Conceptualization, Methodology, Project administration **Saeid Nahavandi:** Conceptualization, Methodology, Project administration **Abdoulmohammad Gholamzadeh Chofreh:** Formal analysis, Methodology, Writing - original draft **Feybi Ariani Goni:** Formal analysis, Methodology, Writing - original draft **Jiří Jaromír Klemeš:** Formal analysis, Methodology, Resources, Project administration, Writing - review & editing.

**Acknowledgements:** Several researchers benefited from the EU supported project Sustainable Process Integration Laboratory—SPIL funded as project No. CZ.02.1.01/0.0/0.0/15_003/0000456, by Czech Republic Operational Programme Research and Development, Education, Priority 1: Strengthening capacity for quality research, based on the SPIL project.